\pdfoutput=1

\documentclass[11pt]{article}

\usepackage{ACL2023}

\usepackage{times}
\usepackage{latexsym}

\usepackage[T1]{fontenc}

\usepackage[utf8]{inputenc}

\usepackage{microtype}

\newcommand{\nlformula}{\textsc{NL2Formula \xspace}}%
\newcommand{\nlformulanospace}{\textsc{NL2Formula}}%
\newcommand{\model}{$f$\textsc{Coder \xspace}}%
\newcommand{\modelnospace}{$f$\textsc{Coder}}%
\usepackage{xcolor}
\def\BibTeX{{\rm B\kern-.05em{\sc i\kern-.025em b}\kern-.08em
    T\kern-.1667em\lower.7ex\hbox{E}\kern-.125emX}}

\usepackage[nounderscore]{syntax}

\usepackage{longfbox}
\usepackage{graphicx}
\usepackage{amsmath}
\usepackage[american]{babel}
\usepackage{microtype}
\usepackage{subcaption}
\usepackage[ruled,,linesnumbered ]{algorithm2e}
\usepackage{tcolorbox}
\usepackage{balance} 
\usepackage{enumitem} 
\usepackage{color}
\usepackage{url}
\usepackage{arydshln}

\usepackage{enumitem}
\setitemize{noitemsep,topsep=0pt,parsep=0pt,partopsep=0pt}

\usepackage{amsmath,bm}
\usepackage{amsmath}
\usepackage{amsfonts}
\usepackage{amssymb}

\usepackage{multirow}

\newcount\DraftStatus  
\definecolor{darkgreen}{rgb}{0,0.5,0} 
\definecolor{purple}{rgb}{1,0,1} 
\definecolor{todocolor}{rgb}{0.9,0.1,0.1} 
\definecolor{fixcolor}{rgb}{0.1,0.7,0.3} 
\definecolor{wycolor}{rgb}{0.9,0.1,0.1} 
\definecolor{hycolor}{rgb}{0.7,0.7,0.3} 

\newcommand{\nbc}[3]{\ifnum\DraftStatus=1
	{\colorbox{#3}{\bfseries\sffamily\scriptsize\textcolor{white}{#1}}}
	{\textcolor{#3}{\sf\small$\blacktriangleright$\emph{#2}$\blacktriangleleft$}}
	\fi}

\newcommand{\draftnote}[2]{\ifnum\DraftStatus=1
	\marginpar{
		\tiny\raggedright
		\hbadness=10000
		\def\baselinestretch{0.8}
		\textcolor{#1}{\textsf{\hspace{0pt}#2}}}
	\fi}

\title{\textsc{NL2Formula}: Generating Spreadsheet Formulas from Natural Language Queries}

\author{
Wei Zhao$^1$\thanks{\hspace{0.5em}
Work was done while Wei Zhao was pursuing a master degree at Huazhong University of Science and Technology, and during an internship at Microsoft.
}~~ Zhitao Hou$^2$~~ Siyuan Wu$^1$~~ Yan Gao$^2$~~ Haoyu Dong$^2$~~ Yao Wan$^{1}$\thanks{\hspace{0.5em}Yao Wan is the corresponding author.} \\
\textbf{Hongyu Zhang}$^3$~~ \textbf{Yulei Sui}$^4$~~ \textbf{Haidong Zhang}$^2$\\
  $^1$Huazhong University of Science and Technology, $^2$Microsoft \\
  $^3$Chongqing University,  $^4$University of New South Wales\\
  \texttt{\{mzhaowei,sy\_wu022,wanyao\}@hust.edu.cn}, \texttt{hyzhang@cqu.edu.cn}\\ 
  \texttt{\{zhith,yan.gao,hadong,haizhang\}@microsoft.com}, \texttt{y.sui@unsw.edu.au} 
}
\begin{document}

\maketitle

\begin{abstract}
Writing formulas on spreadsheets, such as Microsoft Excel and Google Sheets, is a widespread practice among users performing data analysis. However, crafting formulas on spreadsheets remains a tedious and error-prone task for many end-users, particularly when dealing with complex operations. To alleviate the burden associated with writing spreadsheet formulas, this paper introduces a novel benchmark task called \nlformulanospace, with the aim to generate executable formulas that are grounded on a spreadsheet table, given a Natural Language (NL) query as input.
To accomplish this, we construct a comprehensive dataset consisting of 70,799 paired NL queries and corresponding spreadsheet formulas, covering 21,670 tables and 37 types of formula functions.
We realize the \nlformula task by providing a sequence-to-sequence baseline implementation called \modelnospace.
Experimental results validate the effectiveness of \modelnospace, demonstrating its superior performance compared to the baseline models.
Furthermore, we also compare \model with an initial GPT-3.5 model (i.e., \texttt{text-davinci-003}).
Lastly, through in-depth error analysis, we identify potential challenges in the \nlformula task and advocate for further investigation.\footnote{All the experimental data and source code used in this paper are available at \texttt{\url{https://github.com/timetub/NL2Formula}}.
} 
\end{abstract}

\section{Introduction}
It is a widespread practice among users to engage in data analysis by composing formulas within spreadsheet applications such as Microsoft Excel and Google Sheets. 
While spreadsheet formula languages (e.g., Microsoft Excel Formula) are relatively simpler than general-purpose programming languages for data analysis, formulating these formulas on spreadsheets remains burdensome and error-prone for end-users~\cite{gulwani2011automating,cheung2016custodes}.
To address this challenge, numerous approaches and tools (e.g., FlashFill~\cite{gulwani2011automating} and \textsc{SpreadsheetCoder}~\cite{chen2021spreadsheetcoder}) have been proposed to automatically generate spreadsheet formulas.

Building upon substantial progress in spreadsheet formula generation, this paper goes beyond the existing efforts by introducing a novel Natural Language (NL) interface capable of generating spreadsheet formulas from a user's NL query (short for \nlformulanospace). 
We believe that, for the majority of end-users, expressing their intentions in NL is more accessible than working with formulas when performing data analytics on spreadsheets.

\begin{figure*}[t!]
    \centering
    \includegraphics[width=.98\textwidth]{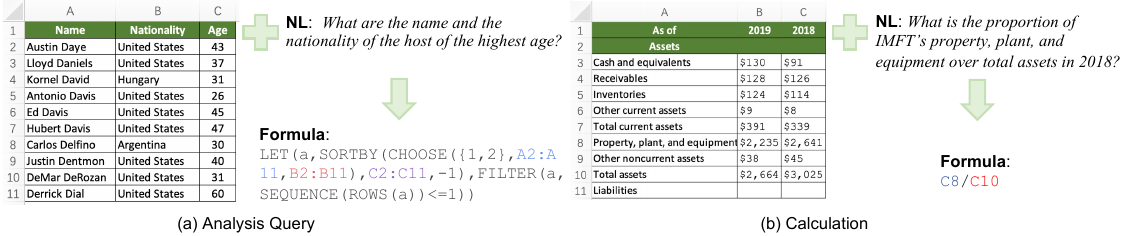}
    \caption{Two running examples from our created dataset for \nlformulanospace. 
    }
    \label{fig:nl2formula example}
\end{figure*}

Figure~\ref{fig:nl2formula example} presents two representative running examples to illustrate the task of \textsc{NL2Formula}. This task involves generating the corresponding spreadsheet formula automatically, given a spreadsheet table and an NL query input from an end-user. The resulting formula is intended for execution in spreadsheet applications, such as Microsoft Excel. In this paper, we focus the spreadsheet application only on Microsoft Excel, where spreadsheet formulas can take on various forms, offering a wide range of possibilities for exploration.
Specifically, we present two primary categories of spreadsheet formulas. The first category is the \texttt{Analysis Query} (Figure~\ref{fig:nl2formula example} (a)), typically comprising Excel formula functions utilized for data analysis. The second category is the \texttt{Calculation} (Figure~\ref{fig:nl2formula example} (b)), consisting of basic numerical operations used for straightforward calculations.

It is important to note that \textsc{NL2Formula} shares similarities with the well-studied task of \textsc{Text2SQL}, which involves translating an NL description into a SQL query grounded on a database table~\cite{yaghmazadeh2017sqlizer, yu2018spider, zhong2017seq2sql}. However, it differs in two fundamental aspects.
\textit{(1) The structure of a spreadsheet table is more flexible than that of a database table.}
Unlike fixed patterns in databases, the metadata (e.g., headers and orientation) of tables in a spreadsheet is optional, and the placement of the table in the layout is highly flexible. This flexibility presents significant challenges when it comes to representing the data.
\textit{(2) The formula is typically expressed by the index of data location.}
In the process of generating formulas, it becomes crucial not only to determine which columns in the table should be selected but also to identify the exact position of the cell containing these values. Additionally, the expression of formulas can change with the placement of the table in the layout.


In this paper, we pioneer the effort to formulate and benchmark the task of \textsc{NL2Formula}. 
One main challenge lies in the lack of  well-labeled data for training.
To tackle this issue, we construct a novel dataset comprising paired NL queries and their corresponding formulas, grounded on specific spreadsheet tables.
As manual labeling would require extensive human effort and time, we opt for an indirect transformation approach using an existing dataset of \textsc{Text2SQL} (i.e., Spider~\cite{yu2018spider}), which is composed of 10,181 NL descriptions along with their corresponding SQL queries.
We devise a set of conversion rules by analyzing the grammar of SQL and Excel formulas.
By applying the formulated conversion rules, we convert SQL queries from the established \textsc{Text2SQL} datasets into formulas suitable for \textsc{NL2Formula}. 
Additionally, to augment the dataset, 
we engage in the manual collection of labeled data following a set of predefined rules.
As a result, we produce a comprehensive dataset comprising 70,799 paired NL queries and formulas, associated with a total of 21,670 tables.

Furthermore, we establish a benchmark for \textsc{NL2Formula}. In this benchmark, we also present \modelnospace, a sequence-to-sequence framework based on the pre-trained language model T5~\cite{raffel2020exploring}. 
As a baseline model, we adapt \textsc{ForTap}~\cite{cheng2021fortap}, originally designed for synthesizing spreadsheet formulas, for comparison. 
We conduct comprehensive experiments and analysis to assess the effectiveness of our proposed \modelnospace. The experimental results demonstrate that \model achieves the highest performance with 70.6\% \textit{Exact Matching Accuracy} and 77.1\% \textit{Accuracy} based on the results of running formulas on a specific engine (i.e., Microsoft Excel).
After conducting a comprehensive analysis of the experimental results, we have identified potential areas for improvement and future directions that warrant further exploration.

In summary, the key contributions of this paper are three-fold. 
(1) 
We are the first to formulate a new task of \textsc{NL2Formula}, that can serve as an interface allowing users to effortlessly translate input NL queries into spreadsheet formulas.
(2) 
We introduce a novel dataset that comprises 70,799 paired NL queries and their corresponding formulas, associated with 21,670 tables.
(3) 
We benchmark several models for the task of \nlformulanospace, including our designed \model that is based on pre-trained T5, as well as \textsc{ForTap}~\cite{cheng2021fortap} that is adapted from \textsc{TuTa}~\cite{wang2021tuta}.

\section{Background and The Problem}


\paragraph{Spreadsheet Formula.} 
Spreadsheets, which are formulated as a two-dimensional grid of cells, play a vital role in our daily lives, especially for data analysis. 
Typically in a spreadsheet, rows are numbered sequentially from top to bottom, beginning at 1, while columns are designated alphabetically from left to right using the base-26 system, with `A' to `Z' as the digits.

\begin{figure}[!h]
    \centering
    \includegraphics[width=.5\textwidth]{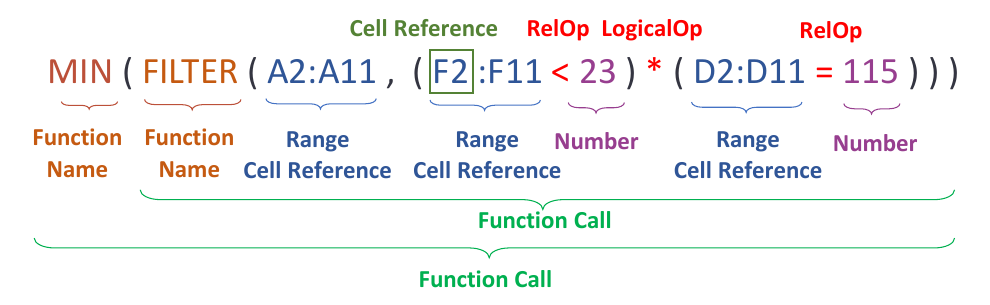}
    \caption{ An example of the Excel formula.}
    \label{fig_grammar_example}
\end{figure}

We can perform various computing, data processing, and operational tasks using pre-defined formulas within the spreadsheet.
In a formula, we can refer to a cell by combining its column and row numbers, as shown by the notation (e.g., \texttt{B2}). Additionally, we have the option to use a range operator ``\texttt{:}'' to create a rectangular range between two cells, with the top-left and bottom-right corners specified. For instance, the formula \texttt{=SUM(A1:B5)} encompasses all cells in columns \texttt{A} and \texttt{B}, ranging from row 1 to row 5.
In general, a formula is composed of constant values, arithmetic operations, function calls, and references to cells.
Formally, the Microsoft Excel formula studied in this paper can be defined by the extended BNF grammar, referred to Appendix~\ref{sec_grammar}.
Figure~\ref{fig_grammar_example} shows a detailed example of the Excel formula.

\paragraph{Problem Statement.}
Let $N$ denote the NL query composed of a sequence of tokens $\{q_1, q_2,\ldots, q_L\}$, and $T$ denote the corresponding tabular context composed of a collection of cells $\{c_1, c_2,\ldots, c_M\}$.
Let $F$ denote the corresponding formula to predict that is denoted a sequence of tokens $\{y_1, y_2,\ldots, y_K\}$. 
Inspired by previous semantic parsing tasks, we formulate the task of \textsc{NL2Formula} as a sequence-to-sequence problem, where the source sequence is the NL query and its tabular contexts, while the target sequence is the formula. 
More specifically, the \textsc{NL2Formula} problem is expressed as follows: given a source NL sequence $N$, as well as the tabular context $T$, the goal is to learn a mapping function $f$ to map the input $\{N, T\}$ into a formula $F$, i.e., $F=f_{\theta}(N; T)$, where $\theta$ is the parameters of model $f$.

\section{\textsc{NL2Formula}: The Dataset}


\subsection{Dataset Construction}
Constructing a paired dataset of NL queries and spreadsheet formulas poses considerable challenges. One approach to tackle this is by inviting experts to generate corresponding NL queries and spreadsheet formulas based on the tabular content. However, this method is time-consuming and labor-intensive, demanding significant human effort. Hence, it drives us to explore alternative ways of indirectly creating the \textsc{NL2Formula} dataset. Fortunately, we discovered a related task called \textsc{Text2SQL}, which has already undergone extensive study. Leveraging this, we develop a converter from the \textsc{Text2SQL} dataset to the \textsc{NL2Formula} dataset. 
The underlying intuition is that both SQL queries and spreadsheet formulas specify the required data in a similar fashion.

\begin{figure}[!t]
    \centering
    \includegraphics[width=.48\textwidth]{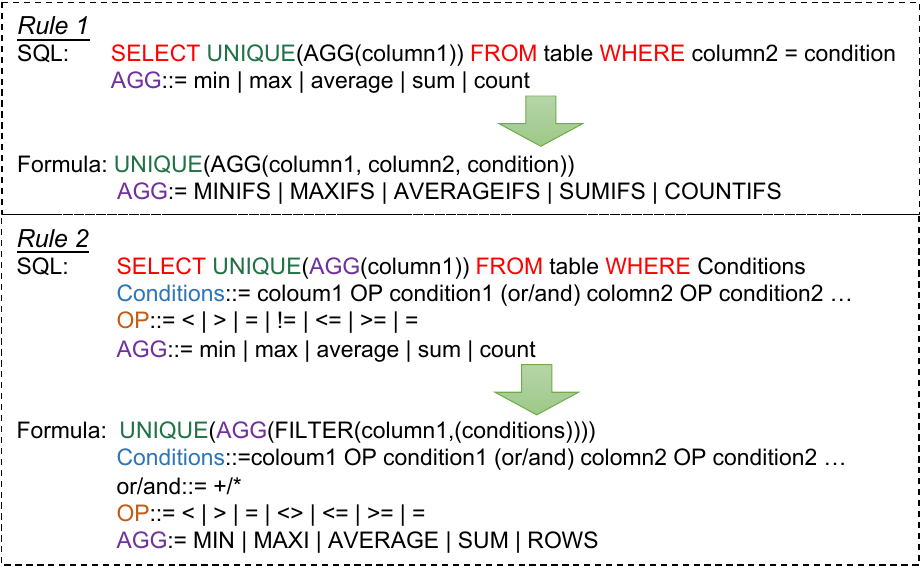}
    \caption{Two simple examples of conversion rules to translate SQL queries into formulas.}
    \label{rules}
\end{figure}

\paragraph{Rule-Based SQL to Formula.}
By analyzing SQL grammar and Excel formula grammar, we manually define several  conversion rules to convert the SQL queries into Excel formulas. 
For example, in certain conditions that necessitate single operations (e.g., \texttt{MAX}) in SQL, we can utilize the corresponding \texttt{MAXIFS} function in a spreadsheet formula. In more intricate scenarios involving multiple conditions and operators in SQL (e.g., \texttt{MIN} and \texttt{AND}), we can replace them with equivalent Excel formulas (e.g., \texttt{MIN} and \texttt{FILTER}).
In situations requiring sorting and combination operations, we need to employ a combination of various Excel formula functions (e.g., \texttt{HSTACK}, \texttt{UNIQUE}, and \texttt{SORT}). 
We present two straightforward examples of conversion rules in Figure~\ref{rules}.

In practice, we primarily utilize two \textsc{Text2SQL} datasets: WikiSQL~\cite{zhong2017seq2sql} and Spider~\cite{yu2018spider}. WikiSQL is an extensive dataset consisting of 80,654 instances of paired NL queries and SQL queries, derived from 24,241 tables sourced from Wikipedia. This dataset exclusively comprises single tables and simple SQL queries. However, our objective is to create a more challenging dataset that encompasses a wider range of formula functions and categories. 
To achieve this, we integrate the Spider dataset, 
with the potential to enhance the diversity of formulas.
Spider is a complex and cross-domain \textsc{Text2SQL} dataset annotated by 11 graduate students. It comprises 10,181 NL queries and 5,693 unique complex SQL queries derived from 200 databases containing multiple tables across 138 different domains. Due to the constraints posed by existing models regarding input data length, we select tables with 3 to 20 rows and 3 to 10 columns. 
As a result, we  obtain approximately 19,789 candidate tables. 


\paragraph{Data Augmentation.}
Based on our investigation, all the formulas converted from \textsc{Text2SQL} are analysis-oriented, commonly referred to as \texttt{Analysis Query}. In other words, these formulas predominantly consist of formula functions such as \texttt{AVERAGE} and \texttt{MAXIFS}. Notably, simple numerical operations such as addition (\texttt{$+$}), subtraction (\texttt{$-$}), multiplication (\texttt{$\times$}), and division (\texttt{$/$}) (also referred to as \texttt{Calculation}) are excluded from the converted formulas. To complement this, we manually augment the data by incorporating a question-answering benchmark named TAT-QA~\cite{zhu2021tat}, which includes numerous numerical operation formulas.


\subsection{Data Statistics and Analysis}
We finally obtain 70,799 pairs of NL queries and spreadsheet formulas, covering 21,670 tables. The tables are randomly split into a training set (75\%), validation set (10\%), and test set (15\%). The basic statistics of each split are shown in Table~\ref{table:dataset_statistic}. The length of a formula is defined by the number of its keywords. We can observe that the average formula length is about 10, indicating the difficulty in predicting these formulas.

To better comprehend the performance of models on various formulas, we categorize the formulas into two groups: \texttt{Analysis Query} and \texttt{Calculation}. 
In particular, \texttt{Analysis Query} formulas encompass 37 types of formula functions, while \texttt{Calculation} formulas consist of addition, subtraction, division, and composition.
Moreover, for \texttt{Analysis Query}, we have tailored the division standards of hardness levels, which are classified into 3 categories: $Simple$, $Medium$, and $Complex$. Specifically, the division standard is based on the number of formula components, selections, and conditions. For instance, we define a formula as $Simple$ if it typically represents a short-length query with 1-2 functions, $Medium$ for 3-4 functions, and any formula with more than 4 functions is considered $Complex$ and falls into the long-length category.

\begin{figure}[!t]
    \centering
    \includegraphics[width=.42\textwidth]{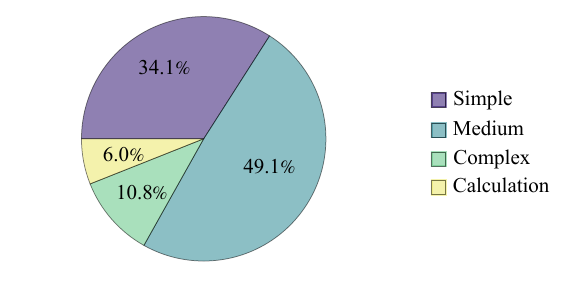}
    \vspace{-3mm}
    \caption{Distribution of formulas in \textsc{Nl2Formula} dataset, including \texttt{Analysis Query} of three hardness levels ($Simple$, $Medium$, $Complex$), and \texttt{Calculation}.}
    \label{fig:hardness_stastic}
\end{figure}

Figure~\ref{fig:hardness_stastic} depicts the hardness distribution of the dataset. It is evident that the majority of formulas consist of medium-level analysis queries, accounting for 49.1\%. 

\begin{table}[!t]
\centering
\caption{Statistics of the \textsc{NL2Formula} dataset.
}
\resizebox{\linewidth}{!}{
\begin{tabular}{l|c|c|c}
\hline
\textbf{Statistics}& \textbf{Train} & \textbf{Val.} & \textbf{Test} \\ \hline
\# of tabular contexts  & 16,791   & 1,743   & 3,136                \\
\# of NL queries& 55,165   & 5,523    & 10,111            \\
Avg. \# of table rows                & 10.8  & 10.8    &10.8             \\
Avg. \# of table columns                & 6.0    & 6.0      & 5.9             \\
Avg. length of NL          & 11.2    & 11.6      &11.4     \\
Avg. length of formula &10.2 & 10.1& 10.0 \\ \hline
\end{tabular}
}
\label{table:dataset_statistic}
\end{table}

\begin{figure*}[t!]
    \centering
    \includegraphics[width=\textwidth]{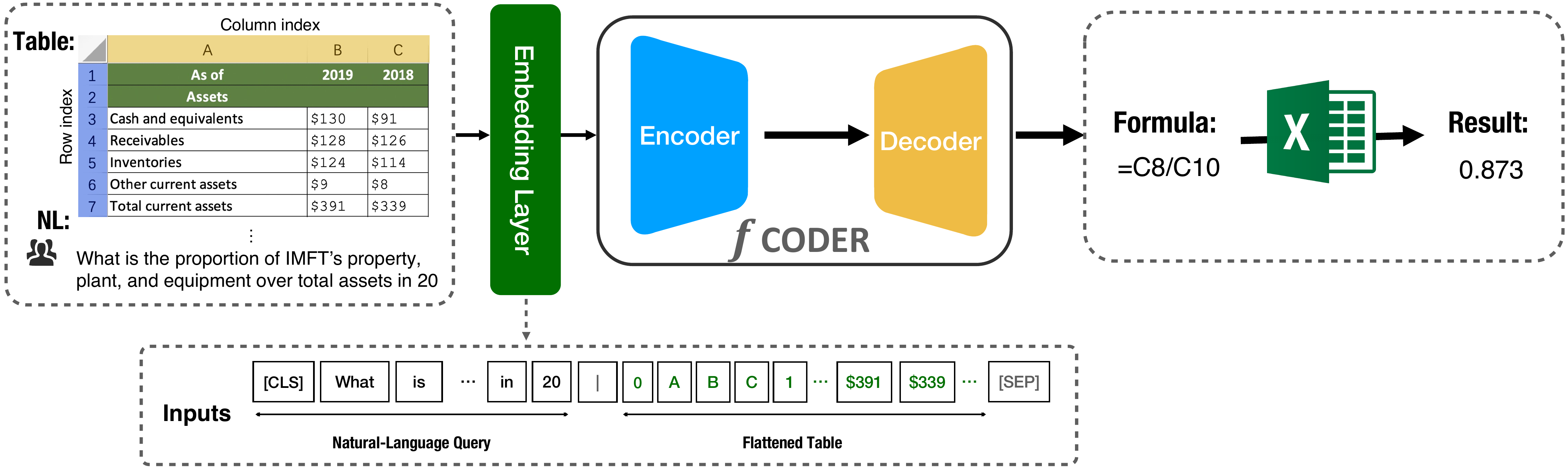}
    \vspace{-4mm}
    \caption{An overview of the \modelnospace, which is a reference framework for \textsc{NL2Formula}. 
    }
    \label{seq2formula_frame}
\end{figure*}

\subsection{Data Quality Assessment}
To ensure the quality of our \textsc{NL2Formula} dataset, we follow a rigorous process. Initially, we randomly sample 5\% of the original data and convert it from SQL queries to formula queries. Subsequently, we input these queries into a spreadsheet to assess their smooth execution.
Based on the execution results, we make necessary adjustments to the conversion rules for formula queries that fail to execute successfully. To guarantee accuracy and reliability, we engage five verifiers with extensive experience in NLP and familiarity with spreadsheet formulas.
Each verifier is tasked with checking and approving 500 pairs of NL queries and formula queries, randomly selected from the dataset. Their expertise ensures meticulous scrutiny of the data.
Finally, in cases where we identify faulty formulas, we verify their formula patterns and search the dataset for all instances of such patterns, making the necessary modifications to rectify the situation.






\section{\model: A Reference Framework}


For the task of \textsc{NL2Formula}, we adopt the encoder-decoder paradigm as the baseline approach. In this paradigm, an encoder network embeds the NL queries and tabular contexts into an embedding vector, while a decoder network generates the formula based on the encoded vector.
Figure~\ref{seq2formula_frame} illustrates the overview of the encoder-decoder framework for \textsc{NL2Formula}.


\paragraph{Input Preparation.}
We represent each table using its column index, row index, and the corresponding content.
Specifically, we work with two types of inputs: an NL query and tabular content. Each input is transformed into a sequence, and subsequently, the two sequences are concatenated. 
We employ a unique symbol \texttt{|} to differentiate between the sequence of NL queries and tabular content. Furthermore, we utilize a specific token \texttt{[CLS]} to mark the inception of the concatenated sequence, resulting in a hybrid representation of the two elements, as follows:
\begin{equation*}
    X=\texttt{[CLS]}, q_1, q_2,\ldots,q_L, \texttt{|}, c_1, c_2,\ldots,c_M\,.
\end{equation*}
For each token $x_i$ in $X$, we begin by encoding it using a word embedding layer, resulting in the token embedding $\mathbf{x}_i^{token}$. Next, we incorporate a positional embedding to account for the position of each token, represented as $\mathbf{x}_i^{position}$. The ultimate embedding of each token for an input sample $X$ is determined as follows:
\begin{equation}
\mathbf{x}_i=\operatorname{Emb}(x_i)=\mathbf{x}_i^{token}+\mathbf{x}_i^{position}\,.
\end{equation}
After processing each token as discussed above, the output sequence is represented by $\mathbf{X} = \operatorname{Emb}(X)$, which serves as the input to the encoder network.

\paragraph{Encoder.}
We input the embedding matrix $\textbf{X}$ into the encoder network, yielding the corresponding output $\mathbf{O}^e$ as follows:
\begin{equation}
\mathbf{O}^e=\operatorname{Encoder}(\mathbf{X})\,.
\end{equation}
Finally, these output embeddings are passed as input to the decoders. 

\begin{table*}[!h]
\renewcommand\arraystretch{1.2}
\centering
\caption{
Overall performance of the \model and baselines on the validation and test datasets, in terms of the EM and ERA metrics.
}

\resizebox{0.9\linewidth}{!}{
\setlength{\tabcolsep}{6pt} 
\begin{tabular}[c]{l|cccc|cc}
\cline{1-7}
{\multirow{3}{*}{\textbf{Models}}} & \multicolumn{4}{c|}{\textbf{Exact Match}} & \multicolumn{2}{c}{\textbf{Execution Result Assessment}} \\ \cline{2-7} 
& \multicolumn{2}{c|}{\textbf{Validation}} & \multicolumn{2}{c|}{\textbf{Test}} & \multicolumn{1}{c|}{\textbf{Validation}} & \multicolumn{1}{c}{\textbf{Test}} \\ \cline{2-7} 
& \textbf{Sketch} & \textbf{Formula} & \textbf{Sketch} & \textbf{Formula}  & \textbf{Formula}  & \textbf{Formula} \\ \hline
\textsc{ForTap}  & - & - & 58.4 & 24.2  & - & - \\ 
GPT3.5 (10-Shot) & - & -& -& 21.4& - &    25.2  \\ 
{$f$\textsc{Coder}}-Small & 97.0 & 65.6 & 96.9 & 65.5  & 71.2  & 70.4 \\ 
{$f$\textsc{Coder}}-Base & 97.4 & 70.5 & 97.2 & 69.4  & 73.3  & 75.0 \\ 
{$f$\textsc{Coder}}-Large & 97.5 & 71.5 & 97.6 & 70.6  & 76.8  & 77.1 \\ \hline
\end{tabular}
}
\label{table:total results}
\end{table*}

\paragraph{Decoder.}
At the $t$-th time step in the decoding process, the operations of the decoder network can be formulated as follows:
\begin{equation}
\mathbf{O}^d_t=\operatorname{Decoder}(\mathbf{O}^e,\operatorname{Emb}(ctx))\,,
\end{equation}
where $\mathbf{O}^d$ is the output of the decoder network, $ctx$ denotes the current partial sequence of the generated formula, i.e., $y_0,\dots,y_{t-1}$, which is also mapped into vector forms via an embedding layer.

We feed the output of the decoder into a Softmax layer, to map the output vector into a probability vector over the whole vocabulary, as follows:
\begin{equation}
    p (y_t |ctx,\mathbf{X}) = \operatorname{Softmax}(\mathbf{W}^d\mathbf{O}^d+\mathbf{b}^d)\,,
\end{equation}
where $\mathbf{W}^d$ and $\mathbf{b}^d$ are the linear layer parameters.



\paragraph{Model Learning.}
To train the \model model, we employ the cross-entropy loss function, as follows:
\begin{equation}
    \mathcal{L} = -\sum_{t=1}^{T}\log p_{\theta}(y_t|ctx,\mathbf{X})\,,
\end{equation}
where $\theta$ denotes all the model parameters, and $T$ is the maximum step of formula generation.

\section{Experimental Evaluation \label{exp_sec}}


\subsection{Benchmarked Models}

\noindent
$\triangleright$ \textbf{{\textsc{ForTap}~\cite{cheng2021fortap}}.}
\textsc{ForTap}, building on \textsc{TuTa}~\cite{wang2021tuta}, extends table pre-training to include spreadsheet formulas for enhanced formula prediction, question answering, and cell type classification. 
We introduce an adaptation of \textsc{ForTap} to \textsc{NL2Formula}, where the task is to predict formulas for a specified cell within a table. We embed the NL query into the table and designate the following row as the target cell. A two-stage LSTM~\cite{hochreiter1997lstm} decoder then processes this integrated data to produce formula sketches and pinpoint reference cells, yielding the target formula.

\noindent
$\triangleright$ \textbf{{GPT-3.5}{~\cite{brown2020language}.}}
With recent advancements in the domain of Large Language Models (LLMs), remarkable breakthroughs have been achieved in the field of NLP~\cite{zhao2023survey,kaddour2023challenges}. In this study, we compare the performance of our proposed methodology with GPT-3.5 on the \textsc{NL2Formula} dataset, utilizing the open-sourced \texttt{text-davinci-003} model.
The prompt template used by GPT-3.5 is referred to Appendix~\ref{sec_prompt}
\begin{table*}[!t]
\centering
\caption{
Experimental results of {$f$\textsc{Coder}} models across different types of formulas, with varying  levels of difficulty on the test dataset.
}
\setlength{\tabcolsep}{2pt} 
\resizebox{\linewidth}{!}{
\begin{tabular}{l|cccc|cccc}
\hline
{\multirow{2}{*}{\textbf{Models}}}  & \multicolumn{4}{c|}{\textbf{Exact Match}} & \multicolumn{4}{c}{\textbf{Execution Result Assessment}} \\ 
\cline{2-9}
    & \textbf{Simple} & \textbf{Medium} & \textbf{Complex} & \textbf{Calculation} & \textbf{Simple} & \textbf{Medium} & \textbf{Complex} & \textbf{Calculation} \\ \hline
GPT3.5 (10-Shot)            & 8.5   & 25.8   & 0.3  & 55.8        & 17.4   & 26.6   & 0.6  & 59.5 \\ 
{$f$\textsc{Coder}}-Small & 39.9   & 73.9    & 54.5   & 62.2   & 58.6   & 82.7    & 56.3  & 64.8          \\
{$f$\textsc{Coder}}-Base  & 44.5   & 76.9    & 53.4   & 71.8   & 63.0   & 87.4    & 56.0   & 74.5        \\
{$f$\textsc{Coder}}-Large & 45.4   & 76.0    & 58.4   & 76.5   & 64.5    & 88.7   & 61.6  & 79.5        \\ \hline
\end{tabular}
}
\label{table:hardless_spilt_result}
\end{table*}

\noindent
$\triangleright$ \textbf{\textit{\modelnospace}.}
We adopt the T5 model~\cite{raffel2020exploring} as the initial implementation of the {$f$\textsc{Coder}} framework. T5 converts all text-based language problems into a text-to-text format and serves as a typical sequence-to-sequence model. 
Some variants of the model are also included in this paper, namely \modelnospace-Small, \modelnospace-Base, and \modelnospace-Large, with parameter sizes of 60M, 220M, and 770M, respectively.

Additionally, we also perform a preliminary comparison between \model and ChatGPT~\cite{gpt_plugins} in the Appendix~\ref{sec_chatgpt}.

\subsection{Evaluation Metrics} 

Inspired by the evaluations in \textsc{Text2SQL}, we also employ two similar metrics: \textit{Exact Match (EM)} and \textit{Execution Results Assessment (ERA)}. Furthermore, we categorize the formulas into two main groups: \texttt{Analysis Query} and \texttt{Calculation}. Additionally, within the \texttt{Analysis Query} category, we further differentiate formulas into three levels, namely, \textit{Simple}, \textit{Medium}, and \textit{Complex}, based on the number of functions they incorporate.

\paragraph{Exact Match (EM).}
The \textit{Exact Match} is a widely recognized metric used to evaluate the performance of models. It demands a flawless match between the model's output formulas and standard formulas, encompassing all its components and table ranges. 
To provide a fine-grained analysis of the model's performance on different granularities of formulas, we present both the \textit{Sketch EM} score and the \textit{Formula EM} score across all models.

\paragraph{Execution Result Assessment (ERA).}
To assess the semantic equivalence of predicted formulas, we also compare their execution results in Microsoft Excel. To streamline this evaluation process, we have developed an automated Python script for large-scale batch execution.

\subsection{Results and Analysis}



\paragraph{Overall Performance}
We begin by analyzing and discussing the overall performance of various models, which includes the baseline \textsc{ForTap}, GPT-3.5, and our proposed {$f$\textsc{Coder}}, on the \textsc{NL2Formula} task. 
Table~\ref{table:total results} presents a comprehensive evaluation of these models on both the validation and test datasets, in terms of the EM (including \textit{Sketch EM} and \textit{Formula EM}) and ERA metrics.

From this table, we can observe a notable performance disparity between the baseline model \textsc{Fortap} and our proposed {$f$\textsc{Coder}} models. The former achieves an EM accuracy of 24.2 on the test set, indicating its struggle to precisely match the ground truth answers. One possible reason is that \textsc{Fortap} is not specifically designed for this task; instead, it focuses on the context of individual cells, neglecting to capture the connections between the entire table and the question. In contrast, the {$f$\textsc{Coder}-Small} model, despite having the smallest number of parameters, significantly outperforms \textsc{ForTap}, achieving an impressive EM accuracy of 65.5 on the test dataset. These results demonstrate the effectiveness of $f$\textsc{Coder} in generating accurate formulas from tabular data.

Furthermore, we can observe that the GPT-3.5 model with a 10-shot in-context learning approach achieves an EM accuracy of 21.4 and an execution results accuracy of 25.2. 
GPT-3.5 model also falls short of matching the performance of the {$f$\textsc{Coder}} series models. This discrepancy could be attributed to the relative simplicity of the current prompt design. Due to the constraints of length of input tokens, we can only provide a prompt consisting of 10 examples at a time, which seems to be insufficient in quantity.

\paragraph{Performance on Varying Hardness} 
We also evaluate the performance of models across both types of formulas, namely, \texttt{Analysis Query} and \texttt{Calculation}, encompassing varying levels of difficulty, as shown in Table~\ref{table:hardless_spilt_result}.
From this table, it is interesting to see that  
our $f$\textsc{Coder} models demonstrate lower performance in the \textit{Simple} level compared to the \textit{Medium} level, in terms of EM accuracy.
Through our human inspection, we have determined that this phenomenon can be ascribed to the fact that the model has a tendency to generate diverse formula queries, primarily stemming from the ambiguity introduced by NL queries.
Furthermore, it is evident that \model attains high performance in the ERA metric. This is attributed to the \model's ability to generate diverse expressions while consistently yielding the correct result.

In comparing the performance of our model with GPT-3.5 utilizing a 10-shot context, it is evident that the GPT-3.5 model exhibits poor performance in generating formulas within the \texttt{Analysis Query} category, highlighting a considerable need for further enhancements. Nonetheless, it is intriguing to observe that the GPT-3.5 model demonstrates a comparable level of proficiency in generating formulas within the \texttt{Calculation} category.


\begin{figure}[!ht] 
    \centering
    \includegraphics[width=.5\textwidth]{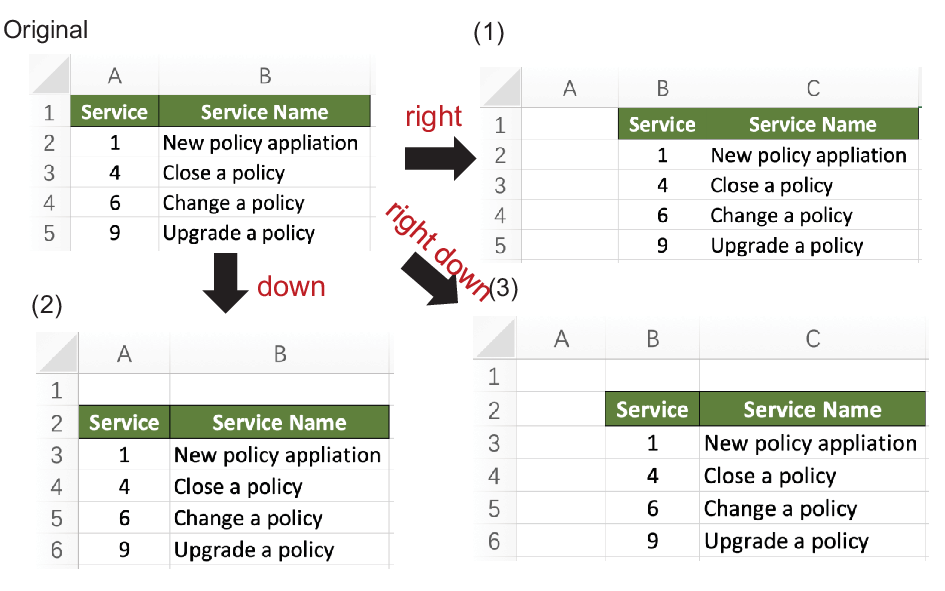}
    \caption{
    An example of a table as well as its three variants of movement in three different directions. 
    }
    \label{fig:indepth_exploration}
\end{figure}
\paragraph{The Impact of Table Position.} 
As previously mentioned, the spreadsheet table is flexible. Therefore, we further explore the performance of the model in generating formulas under different table placements. Specifically, the position of the original tables in our dataset starts from the first row and the column ``\texttt{A}''.
We empirically move these tables in the following three ways, as shown in Figure~\ref{fig:indepth_exploration}:
(1) Moving one column to the right, i.e., the starting position of tables is changed to ``\texttt{B1}''.
(2) Moving one row down, i.e., the starting position of tables is changed to ``\texttt{A2}''.
(3) Moving down and right, i.e., the starting position of tables is changed to ``\texttt{B2}''.
In this scenario, the formulas will also be changed. For example, a formula in the original scenario, \texttt{SORTBY(B2:B5, B2:B5, 1)}, would be transformed to \texttt{SORTBY(C3:C6, C3:C6, 1)} in scenario (3).
Initially, we use the \modelnospace-Base trained in the original position to verify the three scenarios. We explore whether the model can adapt to different table placements in spreadsheets, which were not seen during training. However, the performance of the model is poor, achieving only an average EM accuracy of 6.7\%.
We find that most of the errors are caused by the fact that our model fails to infer the cell index accurately. 

\subsection{Case Study and Error Analysis}
\begin{figure}[h]
    \centering
    \includegraphics[width=.48\textwidth]{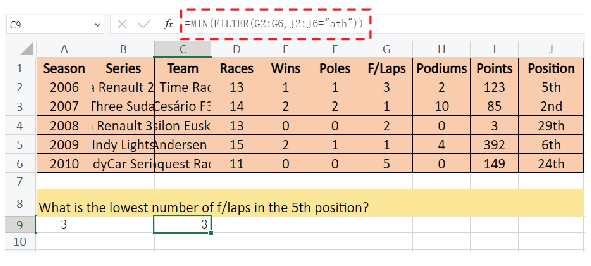}
    \caption{ An example of the prediction formula, which is different from the ground-truth formula but the execution results in the spreadsheet are the same.}
    \label{case_study_1}
\end{figure}

Figure~\ref{case_study_1} presents an illustrative example of the prediction formula, which differs from the golden formula, yet yields identical results when executed in the spreadsheet. The table in \texttt{A1:J6} contains the NL description ``\textit{What is the lowest number of laps in the 5th position?}'' provided in the 8th row. The given golden formula is \texttt{MINIFS(G2:G6, J2:J6, ``5th'')}, and the resulting value after executing this formula in Excel is ``3'', displayed in cell \texttt{A9}. On the other hand, the model prediction formula, \texttt{MIN(FILTER(G2:G6, J2:J6=``5th''))}, produces the same result, which is demonstrated in cell \texttt{C9}.

\begin{figure*}[t!]
    \centering\includegraphics[width=0.98\textwidth]{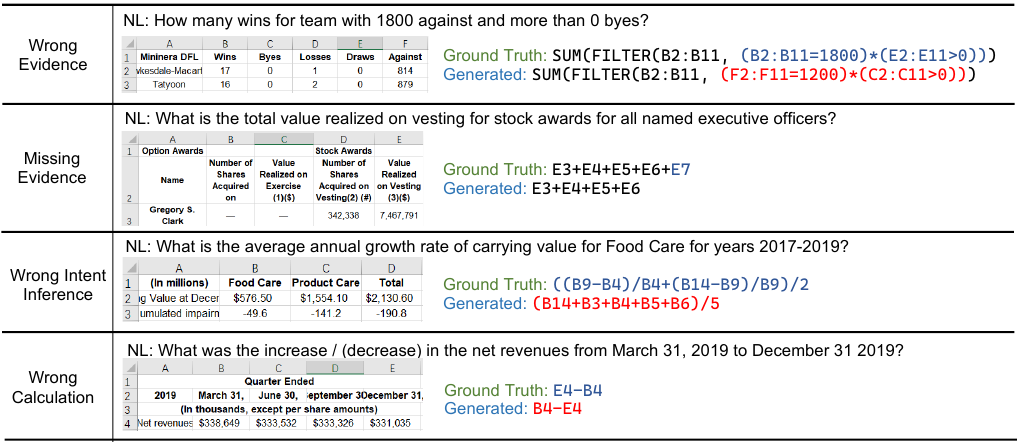}
    \caption{Case studies of error cases. (NL: Natural Language)
    }
    \label{case_study}
\end{figure*}
To gain a comprehensive insight into the effectiveness of our constructed model on \textsc{NL2Formula}, we conduct a detailed examination of the \modelnospace-Large, specifically focusing on instances where errors occur.
We randomly sample 200 error instances from the test dataset (50 per level). We classify them into four categories, as shown in Figure~\ref{case_study}:
(1) Wrong Evidence: The model obtains incorrect supporting evidence or infers the wrong cell index from the table. Additionally, the example of the formula demonstrates the model's failure to identify the correct evidence from the NL query.
(2) Missing Evidence: The model fails to extract complete supporting evidence from the table to arrive at the correct answer.
(3) Wrong Intent Inference: The model is unsuccessful in understanding the intent expressed by the NL query.
(4) Wrong Calculation: The model correctly infers the intention from the NL query and accurately locates the cell index in the table. However, the model fails to compute the answer using the correct evidence.
We find that most of these errors stem from the model's inability to accurately infer or extract the correct evidence from the tables and NL queries.

\section{Related Work}

\paragraph{Semantic Parsing.}
Semantic parsing  is a task to transform NL queries into structured representations that can be understood and processed by machines.
So far, many datasets for semantic parsing have benn built with different query formats, such as ATIS~\cite{1990Evaluation}, Geo-Query~\cite{1996Learning}, and JOBS~\cite{2001Using}. Their output format is logic forms and has been studied extensively~\cite{2016Language, 2014Semantic, 2014Large, 2012Learning, 2007Learning}. In recent years, using SQL queries as programs in semantic parsing is more popular, and many datasets have been built, including Restaurants~\cite{2003Towards}, Academic~\cite{2014Constructing}, Yelp and IMDB~\cite{yaghmazadeh2017sqlizer}, Scholar~\cite{iyer2017learning}, WikiSQL~\cite{zhong2017seq2sql}, Spider~\cite{yu2018spider}, and CoSQL~\cite{yu2019cosql}. 

\paragraph{Formula Synthesis.}
Formula synthesis is a branch of program synthesis that has been studied in many works. 
FlashFill~\cite{gulwani2011automating, gulwani2012spreadsheet} utilizes input-output examples to help end-users automatically synthesize string transformation tasks in spreadsheets. 
Recent studies have explored various neural architectures for learning programs from examples~\cite{kalyan2018neural, parisotto2017neuro}, but they do not consider context-specific information from spreadsheet tables.
\textsc{ForTap}~\cite{cheng2021fortap} and \textsc{SpreedsheetCoder}\cite{chen2021spreadsheetcoder} are the prior approaches for synthesizing spreadsheet formulas from tabular context. 
Our work provides a standardized benchmark for evaluating and comparing future formula generation work, fostering advancement and understanding of the field.


\paragraph{Tabular Data Processing.}
Several studies have pretrained Transformers on tables.
Table-BERT~\cite{chen2019tabfact} linearized tables as sentences so that tables can be directly processed by the pre-trained BERT model. \textsc{TuTa}~\cite{wang2021tuta} is the first effort to pre-train Transformers on variously structured tables. \textsc{ForTap}~\cite{cheng2021fortap} use formulas for numerical-reasoning-aware table pre-training. 
To improve the representation of utterances and tables for neural semantic parsing, several works joined contextual representations of utterances and tables, such as TAPAS~\cite{herzig2020tapas} and \textsc{TaBert}~\cite{yin-etal-2020-tabert}.
Furthermore, \citet{chen2021spreadsheetcoder} introduced \textsc{SpreadsheetCoder}, which leverages machine learning to assist in formula prediction in spreadsheets. 
%

\section{Conclusion}\label{sec_conclusion}
In this paper, we have presented a novel and challenging research problem, \textsc{NL2Formula}, and develop an accompanying dataset that includes spreadsheet tables, NL queries, and formulas. 
We construct a comprehensive dataset consisting of 70,799 paired NL queries and corresponding spreadsheet formulas, covering 21,670 tables and 37 types of formula functions. 
We also realize the \nlformula task by providing a sequence-to-sequence baseline implementation called \modelnospace.
Through in-depth error analysis, we identify potential challenges in the \nlformula task and advocate for further investigation.
We believe that the benchmark developed in this paper can prompote the related research in \nlformulanospace.
\section{Limitations}

There are several limitations  of our research. One is that the formula queries in our \textsc{NL2Formula} dataset are converted from several \textsc{Text2SQL} datasets, resulting in a relatively fixed table structure. Additionally, while we made efforts to include as many formula functions and combinations as possible in our experiments, we have not yet fully covered all types of formula functions, such as the ``\texttt{FIND}'' function used for string queries. In our future work, we aim to expand the range of formula queries by incorporating additional formula functions, specifically targeting a broader array of scenarios. This expansion will include incorporating diverse data samples that utilize functions like ``\texttt{CONCATENATE}'', ``\texttt{LEN}'', and ``\texttt{REPLACE}''. These particular functions are essential for tasks related to data cleaning, preparation, and textual data manipulation. Moreover, we intend to explore the capabilities of models under multi-type tables, including horizontal and vertical tables, to simulate more realistic application scenarios. Furthermore, we aim to investigate situations involving multiple tables under the same spreadsheet.

Another limitation is the maximum length of model input, which is generally 512 characters. Despite controlling the length of rows and columns in the tables in this paper, we observed some errors caused by the model not fully encoding the table.

An additional potential limitation of our approach is the inability to directly execute custom-defined lambda functions in the current Excel environment. The DAX library, with its different grammar from Excel formulas, is used to build formulas and expressions in Excel data models like Power BI, Analysis Services, and Power Pivot. Consequently, we cannot use our execution result metric to measure the performance of custom-defined lambda functions. This limitation may impact the accuracy and comprehensiveness of our evaluation for this specific functionality.

\section*{Acknowledgements}
This work is supported by the National Natural Science Foundation of China under grand No. 62102157.
\bibliography{ref}
\bibliographystyle{acl_natbib}


\appendix
\section{BNF Grammar of Formula}
\label{sec_grammar}
The extended BNF grammar of the Microsoft Excel formula studied in this paper is defined as follows:


\setlength{\grammarindent}{6em} 
\setlength{\grammarparsep}{1pt plus 1pt minus 1pt}
\begin{footnotesize}
\begin{grammar}
    <Formula> ::= "=" <Expr>
    
    <Expr> ::= <Term> \{<AddOp> <Term>\}
    
    <Term> ::= <Factor> \{ <MulOp> <Factor> \}
    
    <Factor> ::= <Number> | <CellReference> | <FunctionCall> |"("<Expr>")"
    
    <CellReference> ::= <ColumnName> <RowNumber>
    
    <ColumnName> ::= <Letter> \{ <Letter> \}
    
    <RowNumber> ::= <Digit> \{ <Digit> \}
    
    <FunctionCall> ::= <FunctionName> "(" [ <ArgumentList> ] ")"
    
    <ArgumentList> ::= <Expr> \{ "," <Expr> \}

    <AddOp> ::= "+" | "-"
    
    <MulOp> ::= "*" | "/"
    
    <RelOp> ::= "<" | ">" | "<=" | ">=" | "=" | "!="
    
    <LogicalOp> ::= "+" | "*"
    
    <FunctionName> ::= [a-zA-Z]+
    
    <Number> ::= <Integer> | <Decimal>
    
    <Integer> ::= <Digit> \{ <Digit> \}
    
    <Decimal> ::= <Integer> "." <Digit> | "." <Digit>
    
    <Letter> ::= [a-zA-Z]
    
    <Digit> ::= [0-9]
\end{grammar}
\end{footnotesize}

\section{Prompt Template Used by GPT-3.5}
\label{sec_prompt}
We utilize a 10-shot in-context learning strategy, where for each new question and table, we dynamically select the Top-10 most similar NL-Formula pair examples from our training set. The similarity is determined based on their BLEU scores~\cite{papineni-etal-2002-bleu}. These selected examples, comprising 10 pairs of NL queries and formulas, are then integrated into a prompt to guide the model in generating its result. We use the following prompt template:
\begin{tcolorbox}
\small
NL: \texttt{[NL description]}\\
Formula: \texttt{[Excel Formula]} 

...(*10)

NL: \texttt{[NL description]}\\
Formula: \texttt{[Excel Formula]} 

NL: \texttt{[NL description]}\\
Formula: \texttt{[to be generated]}
\end{tcolorbox}

\section{Preliminary Comparison to ChatGPT}
\label{sec_chatgpt}

We explore the capabilities of ChatGPT for the task of \textsc{NL2Formula}.
In addition to prompting LLMs to generate formulas (see Sect.~\ref{exp_sec}), we also explore alternative approaches utilizing LLMs for the processing of tabular data.
We leverage Langchain~\cite{Chase_LangChain_2022}, a framework purposefully crafted to harness the potential of LLMs in the realm of application development.
We investigate ChatGPT through two distinct approaches:
(1) Direct Question-Answering~(Direct-QA): We input the complete flattened table directly into the LLMs, prompting it to provide a direct answer to the NL query without any intermediate processing.
(2) Langchain-Agent (Agent): We employ the Langchain CSVAgent workflow, which entails the transformation of the original spreadsheet into a Pandas data frame and the generation of Python code to extract or manipulate data to respond to the NL query. 
We comprehensively evaluate ChatGPT's ability to handle tabular information and respond to NL queries. We randomly select 3,000  samples from the test dataset, which exclusively feature built-in Excel functions and exclude custom-defined lambda functions.
Table~\ref{table:GPT_comparison_result} shows the evaluation results on the \textsc{NL2Formula} dataset.
From this table, we can observe that
ChatGPT exhibits moderate proficiency in processing spreadsheet data. They also unveil limitations in performing basic numerical operations within the \texttt{Calculation} subset, due to their constrained arithmetic and complex reasoning capabilities.
Interestingly, the utilization of ChatGPT with Langchain CSVAgents exhibits notably superior performance when compared to the Direct-QA method. 
This is because the Langchain agent generates Python code for manipulating Dataframes, which closely aligns with the current \textit{Code Interpreter} in handling tabular data.

\begin{table}[!t]
\centering
\caption{Execution results of $f$\textsc{Coder} and ChatGPT, at different levels of hardness.}
\setlength{\tabcolsep}{2pt} 

\resizebox{\linewidth}{!}{
\begin{tabular}{l|cccc|c}
\hline
    & \textbf{Simple} & \textbf{Medium} & \textbf{Complex} & \textbf{Calculation} &\textbf{Overall} \\ \hline
ChatGPT3.5-DirectQA       & 11.5   & 38.9     & 21.1       & 0.8  &  27.7     \\
ChatGPT3.5-Agent & 22.4   & 67.9    & 44.7  & 3.6    &  49.4   \\
\model-Large & 87.0   & 91.6   &  71.1 & 80.5     & 89.1    \\ \hline
\end{tabular}
}
\label{table:GPT_comparison_result}
\vspace{-1em}
\end{table}

\end{document}